\documentclass[conference,letterpaper]{IEEEtran}

\usepackage{amsmath,amssymb}
\usepackage{graphicx}
\usepackage{booktabs}
\usepackage{cite}
\usepackage{array}
\usepackage{xcolor}
\usepackage{hyperref}
\usepackage{url}

\title{Mechanical Interaction Sensing for Robotic Kiwifruit Handling with an Optoelectronic Soft Suction Cup}

\author{
Romina Adhami$^{1}$,
Ebrahim Shahabi$^{2}$,
Alessio Mondini$^{1}$,
Samuel M. Youssef$^{1}$,\\
Cosimo Della Santina$^{2}$,
Gualtiero Volpe$^{3}$,
Barbara Mazzolai$^{1}$\\[0.8em]
{\small $^{1}$Bioinspired Soft Robotics Laboratory,
Istituto Italiano di Tecnologia, Genoa, Italy}\\
{\small $^{2}$Department of Cognitive Robotics,
Delft University of Technology, Delft, The Netherlands}\\
{\small $^{3}$DIBRIS, University of Genoa,
Genoa, Italy}
}

\begin{document}
\pagestyle{empty}
\renewcommand{\IEEEtitletopspaceextra}{23pt}
\maketitle
\thispagestyle{empty}

\begin{abstract}
Robotic handling of delicate fruit benefits from compliant contact and quantitative sensing of the mechanical interaction at the end-effector.
This study investigates an optoelectronic soft suction cup with a kiwifruit-specific bicomponent design for estimating interaction forces during contact with real fruit. 
During known-contact intervals, the optical signals accurately estimate interaction force on fruits excluded from model training: the best shallow network achieves grouped out-of-fold pooled $R^2=0.968\pm0.005$ (MAE $=0.138\,\mathrm{N}$), with a median per-fruit held-out $R^2=0.973$ (MAE $=0.134\,\mathrm{N}$).
A simple monotonic mapping from the mean optical response alone achieves $R^2=0.925\pm0.005$, showing that substantial force-related information is already present in the optical response.
Leave-one-collection-day-out validation remains strong ($R^2=0.955\pm0.031$), supporting robustness across acquisition sessions.
As an exploratory secondary analysis, binary Soft/Firm classification from compression-test reference labels achieves moderate balanced accuracy ($\approx0.705$) under grouped held-out-fruit validation, while finer three-class grading shows limited performance.
\end{abstract}


\section{Introduction}
\IEEEPARstart{P}{ost-harvest} handling of delicate fruit must combine gentle mechanical contact with awareness of the interaction forces that develop at the interface.
Overloading damages tissue; underloading yields unstable grasps.
Kiwifruit is a demanding test case: the surface is curved, compliant, often irregular, and varies strongly among individual fruits.

Soft robotic end-effectors address part of this challenge by conforming to produce and distributing contact pressure~\cite{elfferich2022,ao2025,wang2025jfr,liu2023, elfferich2024berrytwist, montero2024mastering}.
Compliance alone, however, does not provide quantitative feedback on interactions.
Embedding sensors in soft structures is therefore attractive, but the sensor must remain informative when the contacted object is a real fruit rather than a laboratory plate or elastomer sample.

Optical and optoelectronic sensing enables deformation measurement in compliant structures without rigid contact-face transducers~\cite{zhao2016}, while sensorized suction cups combine grasping and sensing~\cite{shahabi2023}.

The optoelectronic suction-cup sensing principle adopted here was previously demonstrated for estimating contact force and direction on controlled surfaces and was integrated into an underwater soft robotic arm~\cite{deldottore2026}.
However, that work did not evaluate the quantitative estimation of the interaction force across biologically variable fruit.
Other studies have investigated tactile force sensing and fruit handling using vision-based~\cite{zhang2026fruittouch} and triboelectric~\cite{sun2026teng} approaches, while piezoresistive sensing has been applied to kiwifruit stiffness estimation~\cite{erukainure2022compag}.
What remains insufficiently characterized is the performance of this optoelectronic suction-cup architecture when adapted for contact with real kiwifruit, particularly its force-estimation accuracy and variability across unseen fruits and acquisition conditions.
This study addresses that gap by evaluating contact-mode force estimation under fruit-grouped and leave-one-collection-day-out validation.

Building on the previously established optoelectronic sensing principle, the contributions of this study are:
\begin{itemize}
    \item Adaptation and attachment-force characterization of a soft suction cup for kiwifruit contact, including geometry and material screening and evaluation of a bicomponent Dragon Skin~10 body with an Ecoflex~00-10 fruit-contacting rim (Figs.~\ref{fig:stage1_screening} and~\ref{fig:stage2_kiwifruit}).

    \item Quantitative interaction-force estimation from optical signals collected during contact without active suction on 70 real kiwifruits, comparing simple and nonlinear models under grouped held-out-fruit validation and examining error variability across fruits, force ranges, loading/unloading phases, and acquisition days (Table~\ref{tab:force} and Figs.~\ref{fig:force_main} and~\ref{fig:robust}).

    \item Exploratory evaluation of fruit firmness classification from the same optical signals using compression-test reference labels and grouped held-out-fruit validation.
\end{itemize}

\begin{figure*}[htpb]
\centering
\includegraphics[width=0.8\textwidth]{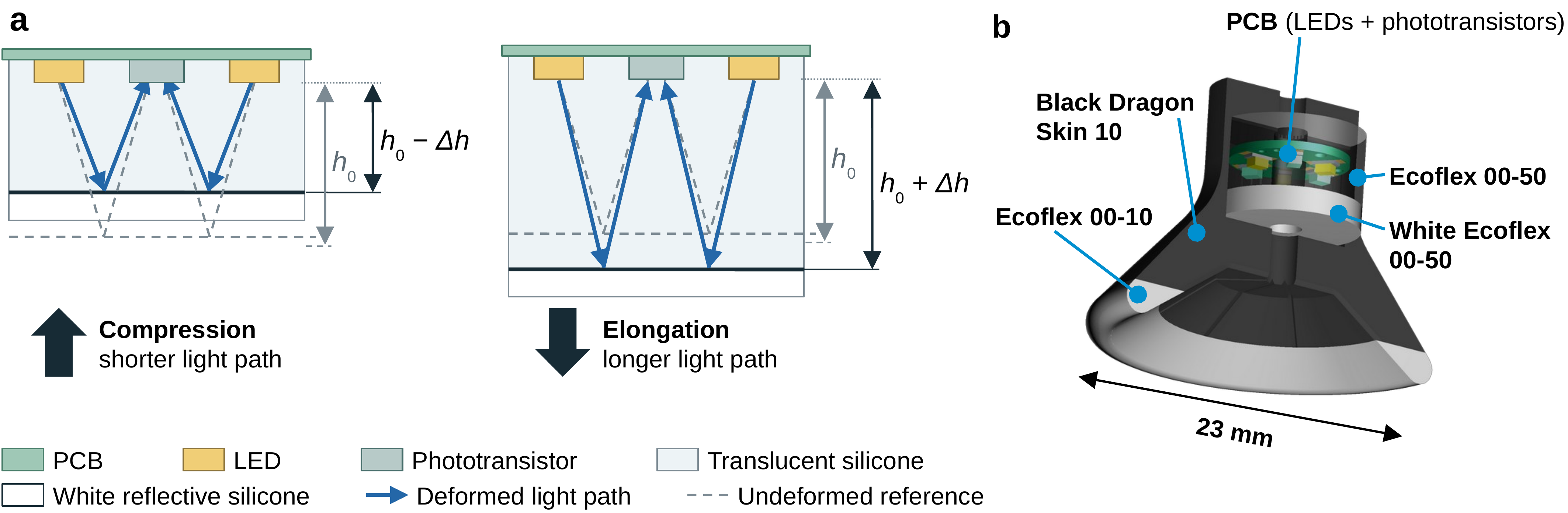}
\caption{Sensorized soft suction-cup design.
(a) Optoelectronic sensing principle illustrating the LED--phototransistor arrangement and deformation-dependent changes in optical paths under compression and elongation.
(b) Final cup construction integrating a PCB with three LED--phototransistor pairs, translucent Ecoflex~00-50 for light transmission, and white reflective Ecoflex~00-50 around the PCB, within a Dragon Skin~10 structural body and a softer Ecoflex~00-10 fruit-contacting rim (base diameter: $23\,\mathrm{mm}$).}
\label{fig:cup}
\end{figure*}

\section{Related Work}

Soft grippers enable compliant fruit handling, while tactile feedback supports damage-aware manipulation~\cite{elfferich2022,ao2025,liu2023,elfferich2024berrytwist,wang2025jfr,mandil2023}. Mo~\emph{et~al.} demonstrated force estimation using stretchable optoelectronic sensing in particle-jamming grippers and qualitative fruit grasping~\cite{mo2024sowa}, while Mao~\emph{et~al.} reported multimodal fiber-optic sensing and delicate fruit grasping~\cite{mao2024finger}. Several reported evaluations use controlled fixtures or material proxies; grouped generalization to unseen biological fruit during multi-day robotic campaigns remains less common.

Sensorized suction cups combine compliant grasping with sensing. Shahabi~\emph{et~al.} used microfluidic sensing in octopus-inspired suction cups to classify Shore-hardness material blocks and estimate contact angle and direction, using synthetic samples~\cite{shahabi2023}. Van Veggel~\emph{et~al.} demonstrated camera-based tactile sensing for closed-loop suction control on rigid objects~\cite{vanveggel2025}. 
Reference~\cite{deldottore2026} reported the optoelectronic suction-cup sensing principle adopted here, using three LED--phototransistor pairs to measure contact-induced deformation and estimate interaction force and direction on controlled surfaces. The sensorized cups were subsequently integrated into an underwater soft robotic arm. Optical soft sensors more broadly encode deformation through intensity modulation in compliant structures~\cite{zhao2016}.

Estimating interaction force differs from inferring intrinsic fruit firmness. Recent tactile-force studies have investigated generalization to unseen objects using vision-transformer regressors~\cite{shahidzadeh2024faf}, calibration transfer across optical sensor variants through domain adaptation~\cite{chen2024dda}, and force tracking on held-out geometries using data-efficient tactile-array calibration~\cite{chelly2025tactile}. However, force estimation during repeated robotic interactions with biologically variable fruit remains underexplored.

Tactile sensing has also been applied to fruit handling and quality assessment. Zhang~\emph{et~al.} (FruitTouch) combined motor-current and vision-based tactile sensing for normal-force estimation, slip detection, softness ranking on silicone replicas, and field harvesting of berries~\cite{zhang2026fruittouch}. Sun~\emph{et~al.} investigated triboelectric sensing for gripping-force and firmness assessment, including kiwifruit measurements consistent with a GY-4 firmness tester~\cite{sun2026teng}. Liao~\emph{et~al.} used vision-based tactile sensing to track hardness changes during mango and kiwifruit softening~\cite{liao2025hardness}.

Kiwifruit firmness has also been investigated using piezoresistive probes. Erukainure~\emph{et~al.} initially evaluated six fruits using elastic-modulus reference measurements~\cite{erukainure2022aim}, followed by a study comparing 63 machine-learning regressors on 1400 measurements~\cite{erukainure2022compag}. Although some models showed limited generalization, validation used measurement-level folds rather than holding out entire fruits. Penetrometer and nondestructive compression methods provide reference approaches for fruit mechanical characterization~\cite{abbott1998,schotsmans2005,sneddon2024}.

Compared with the earlier optoelectronic suction-cup study~\cite{deldottore2026}, the present work retains the established sensing principle but adapts the cup for kiwifruit contact and evaluates quantitative force estimation across unseen fruits and acquisition days.
It complements existing fruit-tactile research~\cite{zhang2026fruittouch,sun2026teng,erukainure2022compag,liao2025hardness} by focusing on the accuracy and variability of interaction-force estimates obtained from this sensing architecture, rather than primarily on harvesting control or intrinsic fruit-property assessment.
Firmness classification is investigated separately as an exploratory secondary outcome.

\section{Materials and Methods}

\subsection{Sensorized Soft Suction Cup}
\label{sec:cup}

A miniaturized circular board embedded in the suction-cup stalk integrates three LEDs spaced at $120^{\circ}$ with three phototransistors. As illustrated in Fig.~\ref{fig:cup}(a), deformation of the compliant sensing structure alters the internal light distribution, producing variations in the three phototransistor signals, $S_1$, $S_2$, and $S_3$. Differential LED-on/LED-off acquisition reduces ambient-light effects.

Adapting this sensing principle to robotic kiwifruit handling required balancing attachment strength with the compliance of the fruit-contacting interface. The cup needed to provide sufficient attachment force for suction-based manipulation while accommodating the curved, irregular, and hairy surface of the fruit. Its geometry and materials were therefore selected through a two-stage characterization process, combining attachment-force screening on reference surfaces with evaluation of the fruit-contact design on real kiwifruit. Fig.~\ref{fig:cup}(b) illustrates the resulting sensorized cup, integrating Ecoflex~00-50 optical sensing components within a Dragon Skin~10 structural body and an Ecoflex~00-10 fruit-contacting rim.

\begin{figure}[t]
    \centering
    \includegraphics[width=\columnwidth]{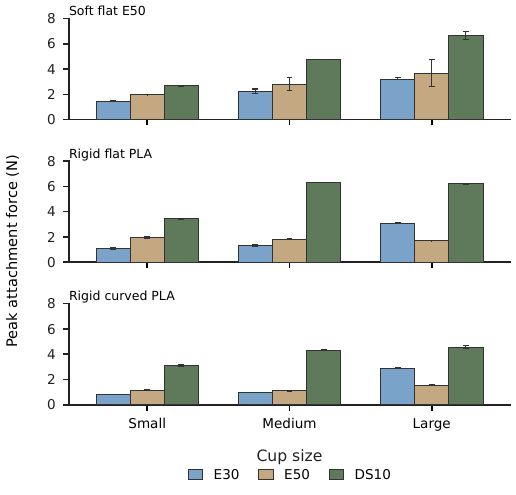}
    \caption{Geometry and material screening of Small, Medium, and Large suction cups made from E30, E50, and DS10 on three reference surfaces: soft flat E50, rigid flat PLA, and rigid curved PLA. Bars indicate mean peak attachment force; error bars show sample standard deviation across repeated pull-off trials within each recording ($n=3$--$7$).}
    \label{fig:stage1_screening}
\end{figure}

\paragraph{Stage~1 --- geometry and material screening}
Candidate cup geometries were initially screened qualitatively for conformity and seal formation on curved kiwifruit surfaces. Attachment performance was then characterized across cup sizes and silicone materials on three reference surfaces: soft flat Ecoflex~00-50, rigid flat PLA, and rigid curved PLA (Fig.~\ref{fig:stage1_screening}). Small, Medium, and Large configurations, with base diameters of $15$, $19$, and $23\,\mathrm{mm}$, respectively, were fabricated from Ecoflex~00-30 (E30), Ecoflex~00-50 (E50), and Dragon Skin~10 (DS10). These materials were selected to investigate attachment performance across a range of silicone hardnesses, from softer Ecoflex formulations to the comparatively harder DS10.

Mean peak attachment force varied with cup size, material, and surface. Overall, DS10 achieved higher mean peak attachment forces than E30 and E50 across the tested conditions. Among the DS10 configurations, the Large cup achieved the highest force on the soft and curved reference surfaces, whereas Medium performed best on the rigid reference. Considering the intended application to curved fruit, the Large DS10 configuration was retained as the structural basis for the kiwifruit-specific design.

\paragraph{Stage~2 --- balancing attachment strength and fruit-contact compliance}
The Stage~1 results identified the Large DS10 configuration as a promising design for achieving high attachment forces. However, handling delicate kiwifruit requires not only sufficient attachment strength but also gentle contact to minimize the risk of fruit damage. In particular, the cup should accommodate the curved, irregular, and hairy surface of the fruit while avoiding excessive localized contact pressure. A thin Ecoflex~00-10 (E10) rim was therefore added to the DS10 body to promote surface conformity and gentler contact while aiming to retain structural support and attachment capability.

\begin{figure}[t]
\centering
\includegraphics[width=0.9\columnwidth]{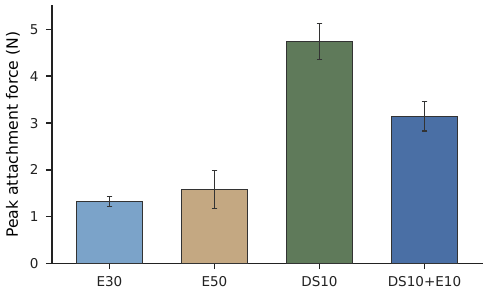}
\caption{Peak attachment force on kiwifruit for four Large suction-cup configurations: monolithic E30, E50, and DS10, and bicomponent DS10+E10. Bars indicate mean values; error bars show sample standard deviation across seven repeated pull-off trials within each recording ($n=7$ per configuration).}
\label{fig:stage2_kiwifruit}
\end{figure}

To assess whether the addition of the softer rim preserved attachment capability for suction-based grasping, the DS10+E10 configuration was evaluated alongside monolithic E30, E50, and DS10 cups of the same Large geometry, this time directly on real kiwifruit (Fig.~\ref{fig:stage2_kiwifruit}). Monolithic DS10 achieved the highest mean peak attachment force ($4.74\,\mathrm{N}$). The bicomponent DS10+E10 cup achieved $3.14\,\mathrm{N}$, exceeding the forces measured for monolithic E30 ($1.32\,\mathrm{N}$) and E50 ($1.57\,\mathrm{N}$).

Although the softer rim reduced peak attachment force relative to monolithic DS10, DS10+E10 retained sufficient attachment capability for the demonstrated whole-fruit grasp. The bicomponent DS10+E10 design was therefore selected for subsequent robotic experiments as a balance between measured attachment performance, structural support, and intended gentle fruit contact.

The selected DS10+E10 cup was subsequently demonstrated in a representative robotic suction-grasping sequence on a whole kiwifruit (Fig.~\ref{fig:grasp_seq}). Suction was generated using a miniature DC vacuum pump (VN-T1, YWBL-WH) at a vacuum pressure of $81\,\mathrm{kPa}$.

\subsection{Robotic Experimental Setup}
\label{sec:setup}
\begin{figure}[!b]
\centering
\includegraphics[width=\columnwidth]{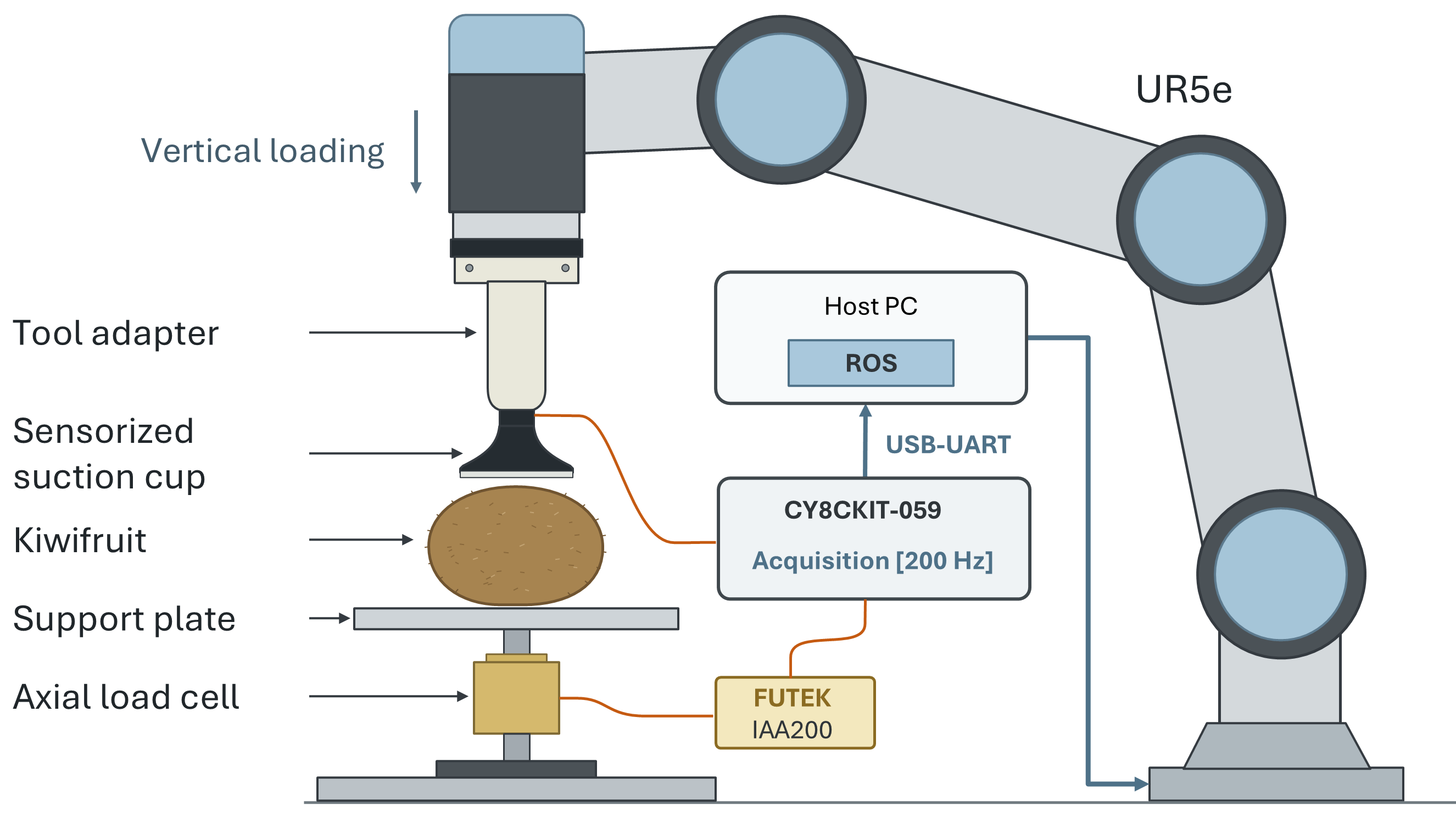}
\caption{Robotic experimental setup for interaction-force data collection. The UR5e manipulator applies vertical loading through the sensorized soft suction-cup end-effector, while an axial load cell beneath the support plate provides reference force measurements. Optical sensor signals and load-cell measurements are acquired through the respective interfaces and recorded on a host PC. A reference object was used for setup and positioning before being replaced by a kiwifruit for experimental data collection.}
\label{fig:setup}
\end{figure}

Kiwifruit interactions were collected using a Universal Robots UR5e manipulator carrying the sensorized suction cup on a 3D-printed mount (Fig.~\ref{fig:setup}). Unlike controlled benchtop loading, this robotic setup enabled repeated interactions with real fruit while incorporating variations in fruit geometry, surface characteristics, contact location, and cup deformation, bringing the evaluation closer to automated fruit-handling applications.

Arm motion was coordinated through a ROS-based interface. During each trial, the end-effector approached the fruit at approximately $100\,\mathrm{mm/s}$. An external miniature S-beam load cell (LSB200, FUTEK Advanced Sensor Technology, USA; 5 lb) was mounted beneath the fruit platform to measure the normal interaction force. Before testing each fruit, the load-cell reading was zeroed with the fruit positioned on the platform, compensating for its static weight so that subsequent measurements reflected the normal interaction force applied by the suction cup during contact.

The load-cell signal was monitored during the commanded downward motion and used as a force-based stopping criterion. The external load cell was used instead of the force estimate provided by the UR5e to obtain a direct and independent measurement of the normal interaction force at the fruit--cup interface. When the measured force reached a prescribed nominal setpoint selected from $\{1.0,1.5,2.0,2.5,3.0,3.5\}\,\mathrm{N}$, a stop command was issued for the downward motion, and the arm subsequently reversed upward to its starting position. The interaction force exhibited a transient overshoot before reversal (median approximately $0.5$--$0.8\,\mathrm{N}$, depending on the setpoint). The load-cell signal also provided the reference force for sensor characterization and model validation.

\begin{figure*}[t]
\centering
\includegraphics[width=0.9\textwidth]{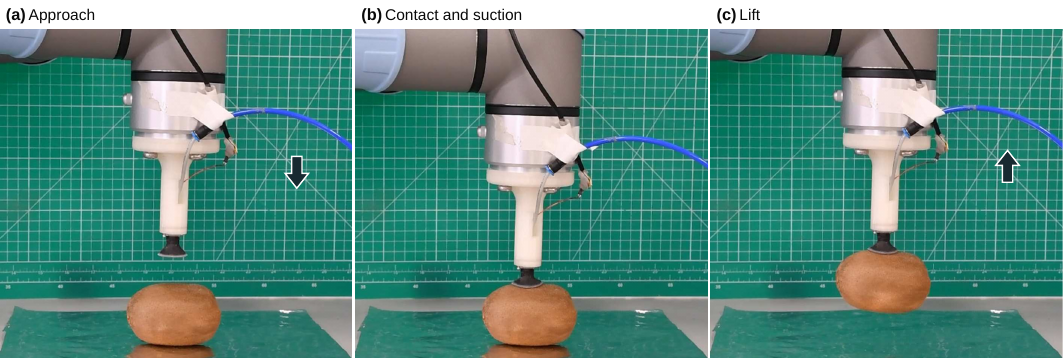}
\caption{Representative suction-based grasping sequence of the selected sensorized suction cup on a whole kiwifruit.
(a)~Approach to the fruit surface.
(b)~Suction-cup contact with the kiwifruit and suction activation.
(c)~Lifting of the fruit under suction.}
\label{fig:grasp_seq}
\end{figure*}

The load-cell bridge signal was conditioned using a FUTEK IAA200 strain-gauge amplifier. The conditioned force signal and three optical channels ($S_1$, $S_2$, and $S_3$) were acquired synchronously at $200\,\mathrm{Hz}$ using a CY8CKIT-059 PSoC 5LP Prototyping Kit (Infineon Technologies), with the load-cell output connected to an analog input channel. Signals were transmitted to the host PC through USB-UART and integrated into the ROS-based acquisition framework. Ambient-light rejection is described in Sec.~\ref{sec:preprocess}.
Force-estimation experiments were conducted in contact mode without active suction. Suction-based grasping was demonstrated separately (Fig.~\ref{fig:grasp_seq}), providing a basis for future integration of interaction-force sensing and grasping in robotic fruit-handling tasks.

\subsection{Kiwifruit Dataset and Experimental Protocol}
\label{sec:dataset}
Data were collected over four purchase/collection-day batches under the same robotic setup.
All quantitative results reported here use the analyzed cohort below. The analyzed force-estimation cohort is summarized in Table~\ref{tab:cohort}.

\begin{table}[t]
\centering
\caption{Analyzed cohort for force estimation}
\label{tab:cohort}
\begin{tabular}{lc}
\toprule
Quantity & Value \\
\midrule
Fruits & 70 \\
Trials & 1509 \\
Nominal force setpoints (N) & 1.0--3.5 \\
Contact $|F|$ range analysed (N) & 0.20--4.81 \\
Collection-day batches & 4 \\
Repetitions per fruit (median/mean) & 16.5 / 21.6 \\
\bottomrule
\end{tabular}
\end{table}

Trials for which physical fruit identity could not be reliably established were excluded from the final analysis. Reference compression tests were performed using a Zwick/Roell Z005 universal testing machine equipped with a $100\,\mathrm{N}$ load cell.
A custom 3D-printed probe was mounted on the testing machine and used to compress the kiwifruit surface.
The probe consisted of a cylindrical indenter, $30\,\mathrm{mm}$ in length and $8\,\mathrm{mm}$ in diameter, terminating in a flat circular contact surface.
During each test, the indenter was driven normal to the fruit surface under position control at a constant speed of $1\,\mathrm{mm/s}$, with a maximum deformation of $8\,\mathrm{mm}$.
The skin was removed at three locations on each fruit before indentation.
For each indentation, the maximum force ($F_{\max}$) reached during compression was recorded, and the mean $F_{\max}$ across the three tested locations was used as the fruit-level reference firmness value.

\subsection{Optoelectronic Signal Processing}
\label{sec:preprocess}

Unless noted otherwise, $S_1$, $S_2$, and $S_3$ denote the three optoelectronic channel responses after reference correction and expression relative to a short start-of-trial baseline, and $S_{\mathrm{mean}}$ denotes their average.

To reduce the effect of ambient light, the LEDs were alternated between on and off states.
For each of the three channels ($i=1,2,3$), the sensor readout was obtained from the absolute difference between consecutive measurements acquired with opposite LED states:
\begin{equation}
S_i^{\mathrm{refcorr}}(k)
=
\left|S_i^{\mathrm{raw}}(k)-S_i^{\mathrm{raw}}(k-1)\right|.
\end{equation}
This procedure preserves an acquisition rate of $200\,\mathrm{Hz}$ while reducing the contribution of ambient light.

Each trial then subtracts the mean of the first $N_0=15$ samples at trial start:
\begin{equation}
S_i(t)=S_i^{\mathrm{refcorr}}(t)-\overline{S_i^{\mathrm{refcorr}}}_{[0,N_0)},
\end{equation}
\begin{equation}
S_{\mathrm{mean}}(t)=\tfrac{1}{3}\sum_{i=1}^{3}S_i(t).
\end{equation}
Some models additionally use the first difference of the mean response, $S_{\mathrm{mean}}(t)-S_{\mathrm{mean}}(t-1)$, computed from past samples only.

\textbf{Contact definition.}
Contact intervals for this analysis were defined \emph{offline} using the reference load-cell signal: samples with $|F|\ge 0.2\,\mathrm{N}$ were retained.
The load cell, therefore, defined known-contact intervals and provided the force target, but load-cell force was \emph{not} supplied as a predictive model feature.
All force-estimation results therefore apply to interaction force \emph{during known contact}.






\subsection{Force Estimation}
\label{sec:force_methods}

Force estimation was evaluated using the absolute load-cell force, $|F|$, as the reference. To assess the force information captured by the optical response, we considered the mean signal $S_{\mathrm{mean}}$, individual channels ($S_1$, $S_2$, and $S_3$), and their combined three-channel representation. We also examined the mean signal together with its causal first difference to account for temporal variations during loading.

Three regression models of increasing complexity were considered. Isotonic regression was used to establish a monotonic relationship between the mean optical signal and force. Ridge regression ($\alpha=1.0$) provided a linear mapping using the mean signal and its causal first difference, with feature scaling performed within each training fold. A shallow multilayer perceptron (MLP), consisting of one hidden layer with 32 ReLU units, was trained using the same inputs, with early stopping and a maximum of 400 iterations. In addition, a progress-only control model was evaluated based on contact and trajectory progress, without optical measurements.

These models were selected to investigate the force information available in the optical signals rather than to compare machine-learning architectures. Fruit identity, collection day, compression-test firmness, nominal force setpoint, and load-cell measurements were excluded from the predictive inputs.


\begin{figure}[htpb]
\centering
\includegraphics[width=\columnwidth]{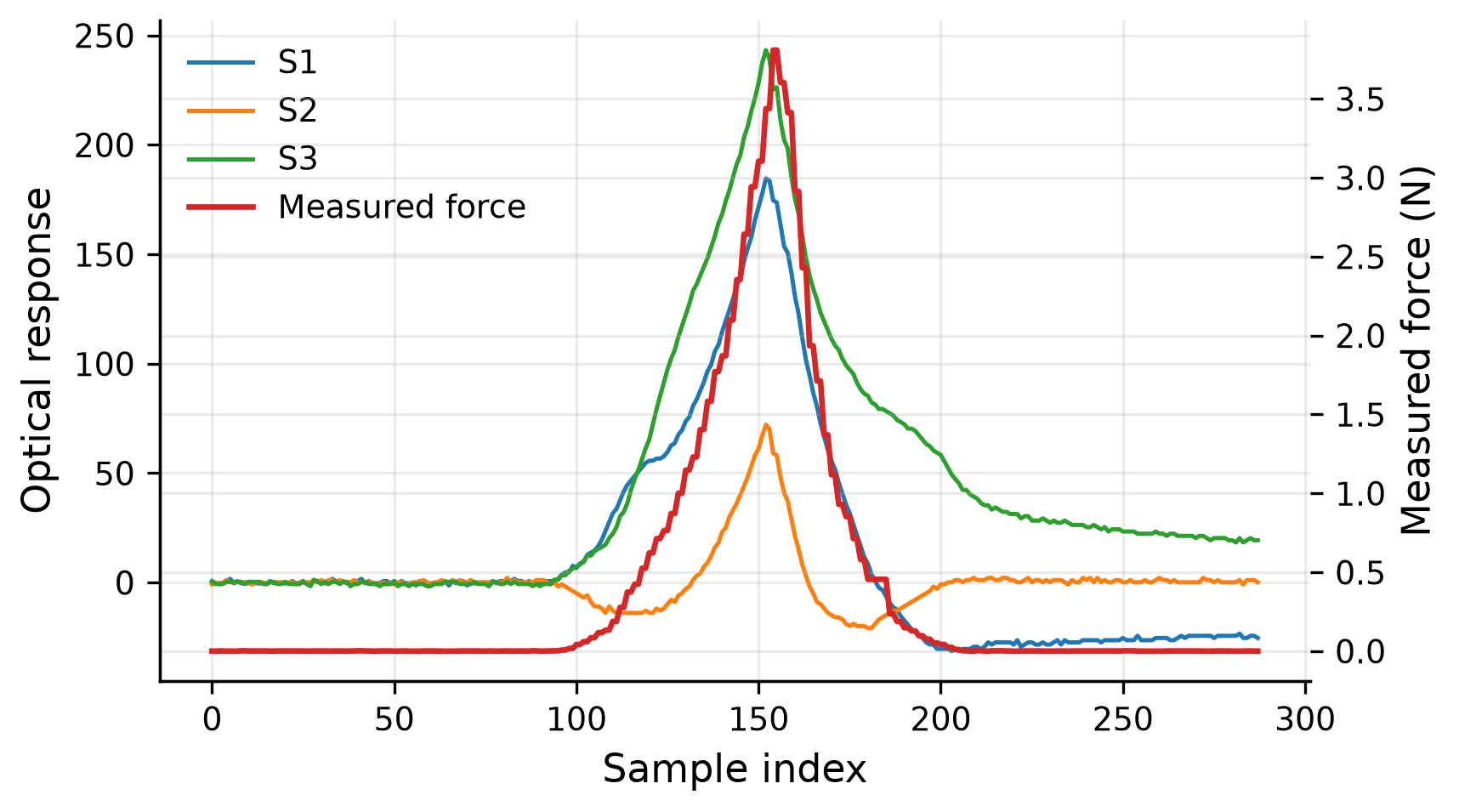}
\caption{Representative kiwifruit interaction showing the three optoelectronic sensor responses (S1--S3) together with the measured load-cell force during contact.}
\label{fig:interaction}
\end{figure}


\subsection{Firmness Assessment}
\label{sec:firmness_methods}

Firmness classification used analysis-defined thresholds, a separate feature-extraction pipeline from force estimation, and a fixed classifier protocol.

\textbf{Cohort and labels.}
The analysis included $675$ alternating-LED trials at the $3.5\,\mathrm{N}$ nominal setpoint from $70$ fruits with reference compression-test measurements. Only trials with experimental quality status PASS, indicating meaningful contact, adequate force-region coverage, and no exclusion or review flags, were retained. Trials with unreliable fruit identity were excluded.

Firmness labels were derived from independent compression tests (Sec.~\ref{sec:dataset}), rather than robotic interaction forces. Two exploratory classification tasks were defined: binary Soft ($<5\,\mathrm{N}$, 47 fruits) versus Firm ($\geq5\,\mathrm{N}$, 23 fruits), and three-class Soft ($<2.5\,\mathrm{N}$, 21 fruits), Medium ($2.5$--$<5\,\mathrm{N}$, 26 fruits), and Firm ($\geq5\,\mathrm{N}$, 23 fruits). These thresholds were defined for this analysis rather than as standardized ripeness categories. Optical interaction descriptors were used to investigate firmness-related differences under comparable loading conditions.

\begin{figure*}[!t]
\centering
\includegraphics[width=0.9\textwidth]{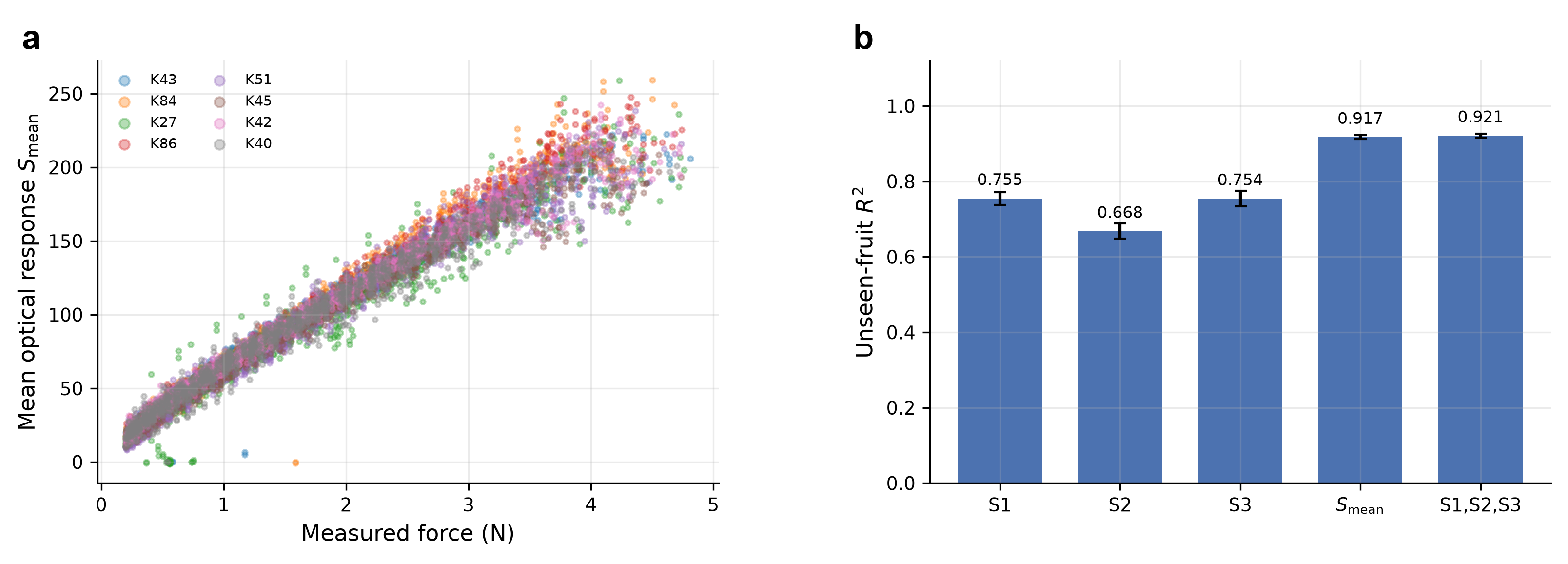}
\caption{Optical response versus interaction force on real kiwifruit.
(a)~Mean optical response $S_{\mathrm{mean}}$ versus measured force during loading for several fruits.
(b)~Force-estimation accuracy on held-out fruits when using individual channels, the mean response, or the joint three-channel input.}
\label{fig:char}
\end{figure*}


\textbf{Feature preprocessing.}
Unlike the start-of-trial baseline used for force estimation (Sec.~\ref{sec:preprocess}), firmness inputs were Phase-2 interaction descriptors extracted from the $3.5\,\mathrm{N}$ trials: LED-offset-corrected channels, a mechanically defined pre-contact region ($|F|<0.2\,\mathrm{N}$), force-based loading segmentation, and per-trial summaries.
The \texttt{E\_combined} feature set comprised $22$ interaction descriptors: channel amplitudes at contact threshold and peak, force-matched responses at $2.0$--$3.0\,\mathrm{N}$, loading slopes, and hysteresis terms.
Load-cell force measurements were used only to define interaction regions and force-matched sampling locations; the resulting predictive features were derived from the optical signals.

\textbf{Models.}
Candidate classifiers included logistic regression, a calibrated linear SVM, and a shallow random forest. For each task, the model achieving the highest trial-level balanced accuracy under grouped held-out-fruit validation was reported. Logistic regression was selected for both binary Soft/Firm and three-class Soft/Medium/Firm classification. Median imputation and standardization were performed within the training folds. Both classifiers used identical hyperparameters: regularization parameter $C=1.0$, balanced class weights, and a maximum of 4000 iterations.

\textbf{Validation.}
Firmness classification used 5-fold stratified cross-validation grouped by fruit identity, repeated with two random seeds. All trials from each fruit remained entirely in the training or test set. Balanced accuracy and macro-F1 were computed over trial-level predictions rather than equally weighting each fruit. Reported values are the mean balanced accuracy and macro-F1 across folds.

\subsection{Evaluation Protocol}
\label{sec:eval}

\textbf{Grouped held-out-fruit validation.}
Force estimation was evaluated using 5-fold cross-validation grouped by fruit identity, repeated with three random seeds. All trials from each fruit remained entirely within either the training or test set, preventing leakage between folds. We report sample-pooled out-of-fold metrics and equal-weight per-fruit summaries (median and mean).

\textbf{Leave-one-collection-day-out (LODO) validation.}
To assess robustness across acquisition sessions, each collection day was held out in turn while training on the remaining days.

\textbf{Additional analyses.}
Force-estimation performance was evaluated separately during loading and unloading and compared against a progress-only control without optical inputs. Firmness classification was evaluated using fruit-grouped splits with trial-level metrics, as described in Sec.~\ref{sec:firmness_methods}.

\section{Results}

\subsection{Sensor Response During Kiwifruit Interaction}
\label{sec:res_response}

The optical responses showed a strong monotonic association with measured interaction force over the loading-to-peak portion of contact (Fig.~\ref{fig:interaction}).
When samples were pooled within each fruit, the mean optical response $S_{\mathrm{mean}}$ yielded a Spearman correlation of $0.989\pm0.005$ (mean $\pm$ SD across fruits).
Individual channels also showed strong but more variable associations with force (S1 $0.922\pm0.030$; S2 $0.897\pm0.049$; S3 $0.940\pm0.026$), consistent with differences in local cup deformation on the curved fruit surface.

\begin{figure*}[t]
\centering
\includegraphics[width=0.9\textwidth]{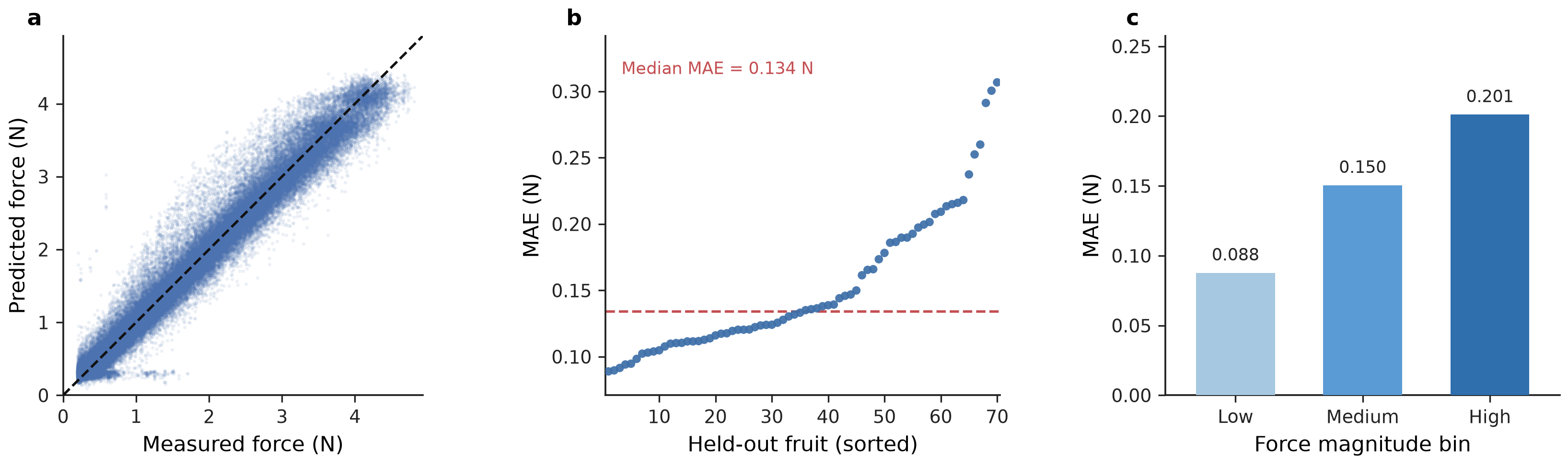}
\caption{Force estimation on held-out kiwifruit.
(a)~Predicted versus measured load-cell force using seed-0 grouped out-of-fold predictions ($n=102{,}640$ contact samples).
(b)~Per-fruit MAE across 70 held-out fruits (sorted); median MAE $=0.134\,\mathrm{N}$.
(c)~MAE by measured force magnitude: low $[0.2,1.0)\,\mathrm{N}$, medium $[1.0,2.5)\,\mathrm{N}$, and high $[2.5,10)\,\mathrm{N}$.}
\label{fig:force_main}
\end{figure*}

\begin{table}[t]
\centering
\caption{Primary force-estimation metrics.}
\label{tab:force}
\footnotesize
\setlength{\tabcolsep}{3pt}
\begin{tabular}{@{}llcc@{}}
\toprule
Validation & Model & $R^2$ & MAE (N) \\
\midrule
Grouped OOF & Isotonic & $0.925\pm0.005$ & $0.236$ \\
Grouped OOF & Ridge & $0.959\pm0.007$ & $0.161$ \\
Grouped OOF & Shallow MLP & $0.968\pm0.005$ & $0.138$ \\
Grouped OOF & Progress-only & $0.212$ & $0.776$ \\
LODO & Shallow MLP & $0.955\pm0.031$ & $0.171\pm0.061$ \\
\bottomrule
\end{tabular}
\end{table}

For the best optical model, the median per-fruit held-out performance is $R^2=0.973$ and MAE $=0.134\,\mathrm{N}$, while the fruit-equal-weight means are $R^2=0.964$ and MAE $=0.152\,\mathrm{N}$ ($N=70$ fruits). Per-fruit MAE ranges from approximately $0.09$ to $0.31\,\mathrm{N}$, revealing variability despite strong overall performance. The fruit-equal-weight mean MAE is higher than the seed-0 sample-pooled MAE ($0.137\,\mathrm{N}$), indicating that pooled metrics do not fully reflect performance across individual fruits.

\subsection{Force-Dependent Response and Channel Behavior}
\label{sec:res_force_dep}
Optical responses increase with measured force during loading across fruits, but fruit-specific gains differ (Fig.~\ref{fig:char}a).
Using only the mean optical response, ridge regression on held-out fruits reaches grouped out-of-fold pooled $R^2\approx 0.917$ (MAE $\approx 0.249\,\mathrm{N}$), nearly matching the joint three-channel model ($R^2\approx 0.921$), while single channels are weaker (Fig.~\ref{fig:char}b).

\subsection{Force Estimation on Held-Out Kiwifruit}
\label{sec:res_force_est}
Figure~\ref{fig:force_main} and Table~\ref{tab:force} summarize the primary force-estimation results.

Under grouped held-out-fruit validation, the shallow neural network achieved the best pooled force-estimation performance ($R^2=0.968\pm0.005$, MAE $=0.138\,\mathrm{N}$), reducing MAE by approximately $14\%$ relative to ridge regression (Table~\ref{tab:force}). Isotonic regression using only the mean optical response also performed strongly, indicating that substantial force-related information is present in this signal. Because isotonic and ridge regression use different input features, the difference in their performance cannot be attributed to temporal information alone.

A progress-only control without optical inputs achieved pooled $R^2\approx0.212$, confirming that optical signals provide information beyond trajectory progress. Leave-one-collection-day-out validation of the best optical model yielded $R^2=0.955\pm0.031$ and MAE $=0.171\pm0.061\,\mathrm{N}$ across four days.

Using the seed-0 grouped out-of-fold predictions of the best model, force-dependent analysis shows that MAE increases with measured force magnitude: $0.088\,\mathrm{N}$ for $[0.2,1.0)\,\mathrm{N}$, $0.150\,\mathrm{N}$ for $[1.0,2.5)\,\mathrm{N}$, and $0.201\,\mathrm{N}$ for $[2.5,10)\,\mathrm{N}$. At higher forces ($\geq2.5\,\mathrm{N}$), the model shows a slight tendency to underestimate force (mean signed error $=-0.048\,\mathrm{N}$). Near the contact threshold ($[0.2,0.5)\,\mathrm{N}$), absolute MAE remains low ($0.074\,\mathrm{N}$), but the median relative error reaches $18.8\%$, indicating reduced relative accuracy at small forces.

\subsection{Loading and Unloading Behavior}
\label{sec:res_load_unload}
\begin{figure*}[t]
\centering
\includegraphics[width=0.9\textwidth]{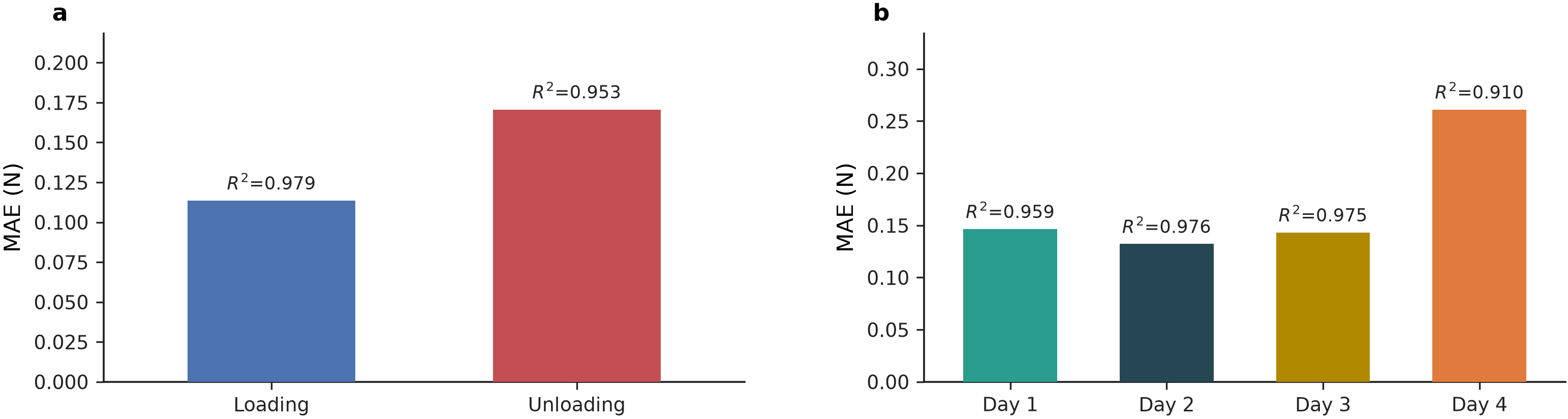}
\caption{Robustness of force estimation.
(a)~MAE during loading (including overshoot) and unloading on held-out fruits ($R^2$ annotated).
(b)~Leave-one-collection-day-out MAE across four collection days ($R^2$ annotated).}
\label{fig:robust}
\end{figure*}

Using the same held-out-fruit predictions, the loading-plus-overshoot subset (loading and overshoot phases) yields pooled $R^2=0.979$ (MAE $=0.114\,\mathrm{N}$), whereas unloading yields $R^2=0.953$ (MAE $=0.171\,\mathrm{N}$) (Fig.~\ref{fig:robust}).
Monotonic loading before overshoot ($R^2=0.974$, MAE $=0.103\,\mathrm{N}$) is similarly high.
Unloading remains informative but weaker, consistent with path dependence.

\subsection{Exploratory Firmness Assessment}
\label{sec:res_firmness}

Binary Soft/Firm classification achieved moderate trial-level performance under grouped held-out-fruit validation (balanced accuracy $\approx0.705$, macro-F1 $\approx0.673$), whereas three-class Soft/Medium/Firm grading was weaker (balanced accuracy $\approx0.490$, macro-F1 $\approx0.479$). Continuous compression-test firmness regression was also not robust as a primary result, indicating that firmness assessment remains exploratory under the present protocol.

\section{Discussion}
\label{sec:discussion}

Accurate force estimation across held-out fruits and collection days indicates that the optoelectronic response remains informative despite variations in fruit geometry and contact conditions within the evaluated protocol. The strong performance of isotonic regression suggests that substantial force-related information is captured by the mean optical signal alone, while the shallow neural network improves accuracy over ridge regression using the same input features. Nevertheless, errors vary across fruits and force magnitudes, particularly near the contact threshold and at higher loads.

Lower accuracy during unloading suggests path-dependent behavior at the cup--fruit interface, potentially arising from material hysteresis or changes in contact conditions during deformation and recovery. Firmness classification was substantially less reliable than force estimation, particularly for three classes. This difference is consistent with the optical signals primarily reflecting sensing-structure deformation under load, whereas compression-test firmness also depends on fruit tissue properties and contact geometry.

Importantly, force estimation was evaluated during known-contact intervals defined using an external load cell, without active suction. Therefore, the results establish contact-mode force-estimation capability rather than autonomous force sensing during suction-based grasping. Integrating contact detection and force estimation into suction-based manipulation remains necessary for closed-loop robotic handling.

\section{Conclusion and Future Work}
This study experimentally evaluated a kiwifruit-specific bicomponent soft suction cup incorporating a previously reported optoelectronic sensing principle~\cite{deldottore2026}. During known-contact intervals without active suction, optical signals enabled accurate interaction-force estimation on unseen fruits, with the best shallow neural network achieving grouped out-of-fold pooled $R^2=0.968$. Strong performance using a simple mean-channel mapping and across held-out collection days further supported the robustness of force estimation. In contrast, exploratory Soft/Firm classification showed only moderate discrimination, indicating that firmness assessment remains a more challenging secondary objective.

Future work will address real-time contact detection, extension of the measurable force range, and integration of force sensing with suction-based grasping in a complete robotic manipulation sequence. Systematic investigation of indentation forces across kiwifruit ripeness levels will help determine the required sensor sensitivity and guide material selection. Evaluation under realistic handling and sorting conditions will be needed to assess practical applicability. Together, these findings provide a foundation for integrating interaction-force sensing and suction-based grasping in robotic fruit handling.
\bibliographystyle{IEEEtran}
\bibliography{references}

\end{document}